\documentclass[letterpaper, 10 pt, conference]{ieeeconf}  

\IEEEoverridecommandlockouts                              

\usepackage{graphics} 
\usepackage{epsfig} 
\usepackage{mathptmx} 
\usepackage{times} 
\usepackage{amsmath} 
\usepackage{amssymb}  
\usepackage{booktabs}
\usepackage{algorithm}
\usepackage{algorithmic}
\usepackage{multirow}
\usepackage{mathtools}
\usepackage{diagbox}
\usepackage{graphicx,overpic}
\usepackage[hidelinks]{hyperref} 

\usepackage[dvipsnames]{xcolor}
\title{\LARGE \bf
StratXplore: Strategic Novelty-seeking and Instruction-aligned Exploration for Vision and Language Navigation
}

\author{Muraleekrishna Gopinathan, Jumana Abu-Khalaf, David Suter and Martin Masek 
\thanks{The authors are with Centre for Artificial Intelligence and Machine Learning, School of Science,
        Edith Cowan University, 270 Joondalup Dr, Joondalup, WA 6027, Australia. {\tt\small \{k.gopinathan, j.abukhalaf, m.masek, d.suter\}@ecu.edu.au}}%
}

\begin{document}

\maketitle
\thispagestyle{empty}
\pagestyle{empty}

\begin{abstract}
Embodied navigation requires robots to understand and interact with the environment based on given tasks. Vision-Language Navigation (VLN) is an embodied navigation task, where a robot navigates within a previously seen and unseen environment, based on linguistic instruction and visual inputs. VLN agents need access to both local and global action spaces; former for immediate decision making and the latter for recovering from navigational mistakes. Prior VLN agents rely only on instruction-viewpoint alignment for local and global decision making and back-track to a previously visited viewpoint, if the instruction and its current viewpoint mismatches. These methods are prone to mistakes, due to the complexity of the instruction and partial observability of the environment. We posit that, back-tracking is sub-optimal and agent that is aware of its mistakes can recover efficiently. For optimal recovery, exploration should be extended to unexplored viewpoints (or frontiers). The optimal frontier is a recently observed but unexplored viewpoint that aligns with the instruction and is novel. We introduce a memory-based and mistake-aware path planning strategy for VLN agents, called \textit{StratXplore}, that presents global and local action planning to select the optimal frontier for path correction. The proposed method collects all past actions and viewpoint features during navigation and then  selects the optimal frontier suitable for recovery. Experimental results show this simple yet effective strategy improves the success rate on two VLN datasets with different task complexities.   

\end{abstract}

\section{INTRODUCTION}

Path planning and navigation in previously unseen environments is a challenging and widely studied problem in robotics. Vision-Language Navigation (VLN) is a robotic task that aims to impart  language-conforming path planning capabilities in robots \cite{Huang2010NLAV,Anderson2018R2R}. Current state-of-the-art methods in VLN build and utilise topological representation of the environment for path planning. However, there is a significant gap between how humans and VLN agents navigate in unseen real-world environments \cite{Sigurdsson2023RREx-BoT,Gopinathan2024SAS}. This is attributed to the diversity of the environment and the arbitrariness of human language. In particular, agents performing long-horizon language-following tasks can become perplexed in unseen environments and eventually make mistakes \cite{Wang2021SSM}. In this paper, we address the challenge of path planning in unseen environments and propose a novel strategy for VLN agents to recover from navigational mistakes.

\begin{figure}
    \centering
    \includegraphics[width=0.9\columnwidth]{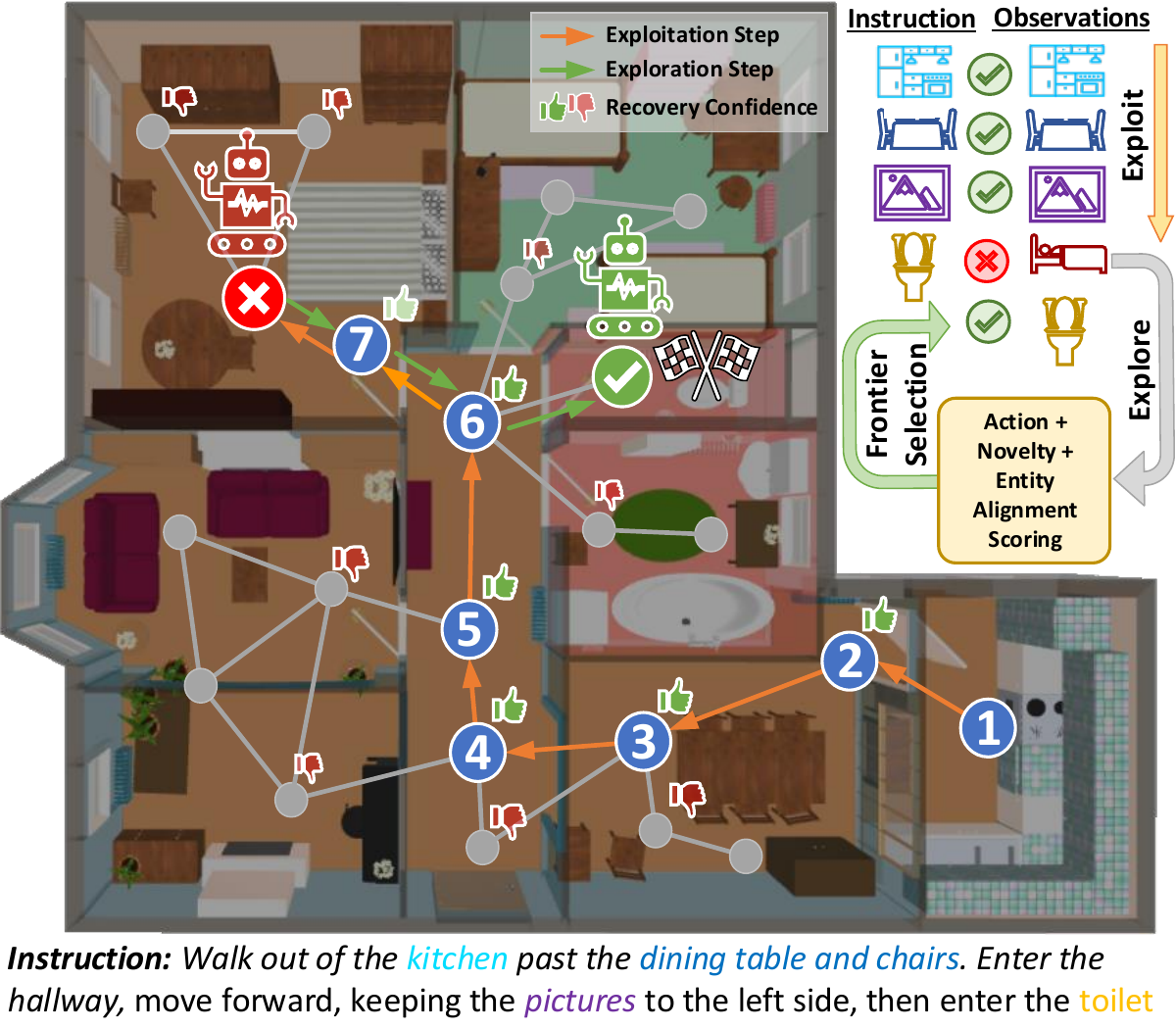}
    \caption{\textbf{Overview}. StratXplore enables an embodied agent to correct its path by exploring frontiers that are both novel and conforms to the given instruction. Here, \textit{exploit} refers to selecting one of the local candidate directions and \textit{explore} considers all unexplored frontiers from the memory.}
    \label{fig:poster}
    \vspace{-1.5em}
\end{figure}

Let us picture a real-world scenario, where a human is given an instruction to ``... move forward, keeping the pictures to the left side, then enter the toilet ..." (Fig. \ref{fig:poster}). Although it may seem straightforward, this can potentially be ambiguous if, for example, the toilet is not visible from the current viewpoint~\includegraphics[height=0.8em]{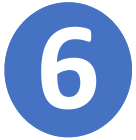}. The human may end up entering the wrong room on the left instead (e.g. bedroom)~\includegraphics[height=0.8em]{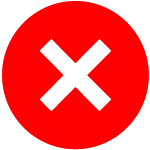}. Typically, when a human makes a navigational mistake, they backtrack and pursue another direction. Similarly, robots can face ambiguity in long-horizon navigational tasks where turn-by-turn instructions are unavailable or the environment is not fully observable. Therefore, in an open-vocabulary real-world setting, the performance of a robot will be heavily dependent on their path planning and error recovery \cite{Boyuan2023OVQSR}. 

Now a natural question is, \textit{How can the agent recover from navigational mistakes?} Traditional strategies applied to object-search problems, suggest that \textit{curiously} seeking novel viewpoints (novelty-seeking) can benefit error recovery and task success \cite{Pathak2017Curiosity}. While enticing, directly applying this method is impractical in VLN task because of its strict need for instruction-path agreement. Hence, exploring for the sake of \textit{curiosity} may result in the agent deviating form the correct path. Instead, hierarchical planners \cite{Irshad2021RoboVLN,Chen2022DUET,An2022BEVBert,Hwang2023MetaExplore} perform back-tracking using dual-scale planning; fine-scale for local planning and coarse-scale for back-tracking. The agent `jumps' to a frontier (unexplored viewpoint) if the planner assigns a higher action probability to that frontier  than to the current candidate directions. These methods have two main limitations; (1) their environment \textit{state} representation, used for planning, is cluttered with previous correct and incorrect visited viewpoints, suppressing the importance of optimal frontiers in decision making (2) these strategies allocate equal significance to all viewpoints, irrespective of how close they are to the goal. We aim to combat the limitations of a hierarchical planner.

The first issue can be addressed by strategically selecting \textit{relevant} frontiers based on their novelty (amount of new information) and correspondence with the given instruction. The second issue can be tackled by prioritising temporally \textit{recent} frontiers over the initial ones. We posit that the \textit{recent} frontiers are more likely to be closer to the goal than initial frontiers making them more significant for recovery. Here, the final selection of potential frontiers is a set of all unexplored viewpoints ranked by decreasing  order of \textit{relevance} based on the recency of the observation.

Our strategic path planning is performed in two steps. Initially, the agent exploits the action decisions from a cross-modal action proposal module (introduced in \S\ref{sec:stratxplore}) while also predicting the agent's confidence in candidate directions (depicted by \includegraphics[height=0.8em]{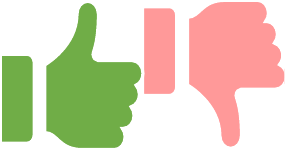} in Fig. \ref{fig:poster}). When the agent learns that it made a mistake based on the confidence scores, it switches to the exploration mode. Here, the confidence score for a direction signifies the likelihood of an agent to reach an instruction-aligned path, if it were pursued. 
During exploration, StratXplore ranks \textit{relevant} frontiers and selects the optimal frontier with the highest rank. Our agent navigates to the optimal frontier via the shortest path to correct the error. Finally, the agent switches to the exploitation mode.

Our contribution in this paper is as follows:
\begin{enumerate}

    \item We propose a novel progress monitoring method that quantifies the agent's confidence in task conformity, if any of the candidate direction is pursued next (\S\ref{sec:stratxplore}). This progress signal is simpler to estimate compared to existing methods.
    \item We introduce a new exploration strategy to select the optimal frontier based on global and local landmark information for recovery. This is determined by the viewpoint temporal recency, viewpoint novelty and instruction-viewpoint correspondence (\S\ref{sec:akfs}).  To the best of our knowledge, this is the first study in VLN on this front. 
    \item We propose an auxiliary learning task that trains the multi-modal planner to identify deviation from an instructed trajectory (\S\ref{sec:pretraining}).
\end{enumerate}

\section{Related Work}
\subsection{Path planning in VLN}
\label{sec:rw-pp}
Path planning is a crucial capability for any navigation agent \cite{Luo2022Stubborn,LozanoPerez1979}. Conventional Vision-and-Language Navigation (VLN) agents are typically constrained to local action space, where choices are limited to the current candidate directions \cite{Chen2021HAMT,Hong2021RecVLNBERT}. Error recovery using these myopic strategies leads to repeated actions, requiring an agent to back-track via each visited step and re-evaluate them. To determine when to back-track, recent error recovery methods \cite{Zhu2022DistPred,Ma2019SelfMonitoring} estimate the progress based on visited viewpoints and back-track if the progress is diminishing. AuxRN \cite{Zhu2020AuxRN} used progress monitoring as a training objective instead. However, these sparse progress signals are harder to estimate during navigation in unseen environments, as the distance to the goal is unknown without pre-exploration. Additionally, another common limitation of these agents is that  only visited viewpoints are stored in their memory, limiting the action space. 

Transformer-based planners \cite{Chen2022DUET,An2022BEVBert}, ameliorate this shortcoming by employing hierarchical decision making on global and local action spaces. The local and global decision contexts are generated independently from two cross-modal transformers and fused later for action prediction. Late fusion makes the global planner oblivious to local planner's decisions and vice versa. Furthermore, this fusion method fails to take advantage of past action scores and the significance of recent observations over initial ones. Our method incentives the agent to select recent viewpoints that align with the instruction both globally and locally.

\subsection{Memory representations in VLN}
\label{sec:rw-memrep}
Memory in robotics encompasses the environmental and navigational \textit{state} representations of an agent which can be used for localisation, scene understanding, question answering, object discovery and instruction following \cite{Wang2021SSM,Gopinathan2021SSU,Kwon2021VGM}. Path planning in a fully explored environment can be reduced to a shortest path problem \cite{Sigurdsson2023RREx-BoT} and an abstract-level memory (comprising of low-detail scene features) is sufficient for the agent. In contrast, a more detailed memory structure is crucial for an agent navigating in unseen environments. Inspired from human memory, robotic memory evolved from storing metric \cite{Elfes1987Sonar} to semantic \cite{An2022BEVBert} information in the form of recurrent \cite{Hong2021RVLNBERT}, topological \cite{Chen2022DUET}, hierarchical \cite{Chen2021HAMT}, and topo-metric \cite{An2022BEVBert} representations. Notable works in visual navigation, VGM \cite{Kwon2021VGM} and WGM \cite{loynd2020WMG}, encode visual graph memory comprising of scene features, but the memory is used solely for localising the agent. This method cannot be directly adopted for the VLN task where the exploration scheme should guarantee instruction-trajectory correspondence. Methods that use memory for path planning such as SSM \cite{Wang2021SSM}, store viewpoints from correct and incorrect actions in the memory. A disproportionate amount of incorrect actions can cause the transformer-based planner to pay attention to incorrect viewpoints, causing the planner to make sub-optimal decisions. Instead, we propose curating the viewpoint features that are added to the memory dependent on their local relevance, temporal-importance, and novelty. Our method, which also uses a graph memory,  employs a buffer that accumulates object-centric features from viewpoints. This is used for identifying unique viewpoints (for novelty) and comparing instruction-viewpoint correspondence with other frontier viewpoints.       

\section{Our Approach}
\label{sec:stratxplore}
\begin{figure*}[ht]
  \centering 
  \begin{overpic}[width=0.8\textwidth,percent]{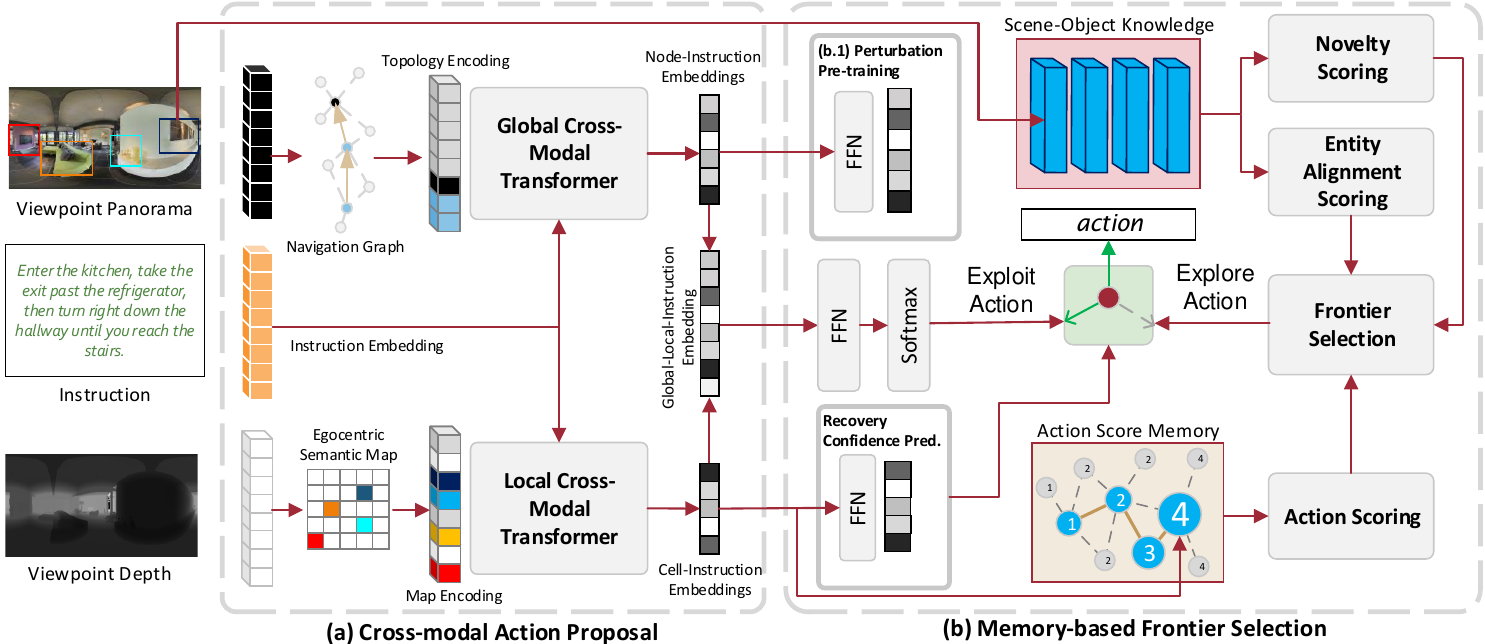}

  \put(2,4){\hyperref[sec:inputs]{\phantom{\rule{52pt}{150pt}}}}

  \put(20,28){\hyperref[sec:topo_enc]{\phantom{\rule{45pt}{50pt}}}}
  \put(18,19){\hyperref[sec:inst_enc]{\phantom{\rule{55pt}{10pt}}}}
  \put(19.5,4){\hyperref[sec:ego_enc]{\phantom{\rule{45pt}{50pt}}}}

  \put(69,31){\hyperref[sec:so_memory]{\phantom{\rule{55pt}{54.5pt}}}}
  \put(69,4){\hyperref[sec:act_memory]{\phantom{\rule{55pt}{50pt}}}}

  \put(54,27.5){\hyperref[sec:pretraining]{\phantom{\rule{44pt}{72pt}}}}
  \put(54,3.5){\hyperref[eq:conf]{\phantom{\rule{40pt}{63pt}}}}

  \put(75.5,17.5){$\scriptstyle{S^{conf}}$}
  \put(44,34){$\scriptstyle{\mathbf{G_t}}$}
  \put(29,25.5){$\scriptstyle\mathbf{N_t}$}
  \put(29,15.5){$\scriptstyle\mathbf{M_t}$}

   \put(67,20){$a_t$}

   \put(65,33){$\scriptstyle{K_{O_t}}$}

  \put(44,10){$\scriptstyle\mathbf{C_t}$}
  \put(31,19.5){$\scriptstyle\mathbf{W}$}

  \put(85,37){\hyperref[sec:novel_score]{\phantom{\rule{50pt}{30pt}}}}
  \put(85,29.5){\hyperref[sec:ea_score]{\phantom{\rule{50pt}{30pt}}}}
  \put(85,18.1){\hyperref[sec:frontierselection]{\phantom{\rule{50pt}{35pt}}}}

\put(71,19.1){\hyperref[sec:frontierselection]{\phantom{\rule{30pt}{30pt}}}}
  
  \put(85,5.5){\hyperref[sec:act_score]{\phantom{\rule{45pt}{30pt}}}}

  \put(54,17){\hyperref[sec:act_prop]{\phantom{\rule{35pt}{46pt}}}}

  \put(31,28.8){\hyperref[eq:trans]{\phantom{\rule{55pt}{47pt}}}}
  \put(31,4.2){\hyperref[eq:trans]{\phantom{\rule{55pt}{47pt}}}}

  \end{overpic}
  \caption{\textbf{Model Architecture of StratXplore.} (a) Fused action proposal from Global and Local Cross-modal transformers (CMT) is used for exploitation (b) When the recovery confidence $S^{conf}$ of current candidates drops below a threshold $c_{thresh}$, the agent chooses to explore. Frontier selector considers the optimal recent-and-novel and instruction-aligned frontier to explore. Blocks are hyperlinked to relevant sections.}
  \label{fig:StratXplore}
      \vspace{-1.5em}
\end{figure*}
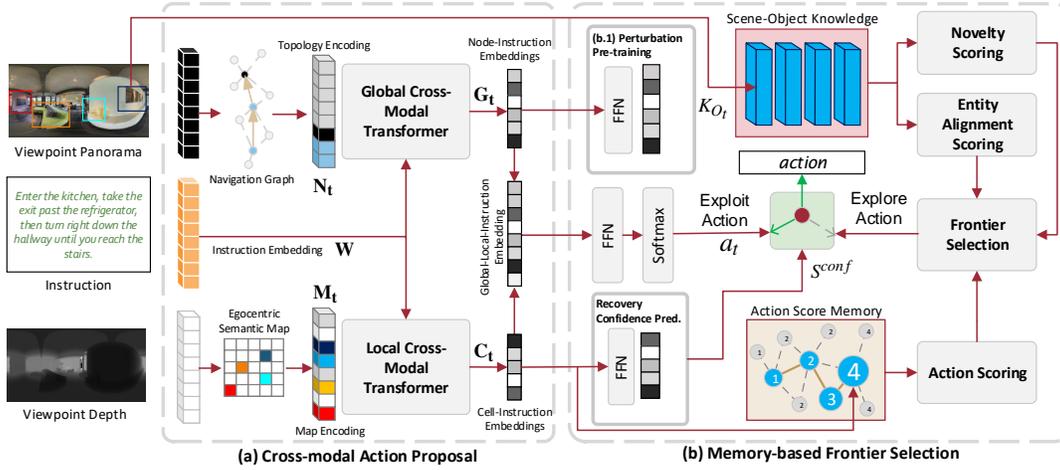
In this section, we introduce the general VLN problem and our strategic exploration scheme for language-guided navigation agents, called StratXplore. The architecture of our method is depicted in Fig. \ref{fig:StratXplore}. 
\subsection{Task Definition}
The Vision-and-Language Navigation (VLN) task integrates natural language processing and visual perception in a pre-defined graph based environment. In VLN, an agent starts at an initial location within  a new indoor environment and follows language instructions to reach the goal. The agent does not have access to the full environment graph during the navigation. At each time step, the agent observes a panoramic viewpoint which comprises 36 single views. The navigable subset of these views is referred to as \textit{candidate} directions. The agent selects one candidate direction to move to the next viewpoint and this process repeats until the agent decides to stop. An ideal agent strictly follows the instruction and stops within 3 meters of the goal location.


\subsection{Inputs}
\label{sec:inputs}
The inputs to the agent are instruction, panoramic viewpoint images, panoramic depth maps, and the agent's world poses (location and orientation) at each step. We use these inputs to generate the following modality encodings. 
\subsubsection{Instruction Encoding}
\label{sec:inst_enc}
The word tokens from the instruction are embedded using an embedding layer. The resulting embedding is summed with the token position embedding and fed to a multi-layer language transformer to obtain the contextual instruction representation \textbf{W}.

\subsubsection{Topology Encoding}
\label{sec:topo_enc}
The panoramic image $\mathbf{O_t}$ of the viewpoint node, observed at each step, is applied to a vision transformer to obtain a viewpoint feature. The pose embedding is $[cos(\theta), \sin(\theta), \cos(\phi), \sin(\phi)]$, where $\theta$ and $\phi$ are relative heading and elevation, respectively. The node embedding of the viewpoint is integrated into the navigation memory, by adding the viewpoint features, pose embedding of the step, and step index. A special stop direction is also included to indicate a stop action. The node embeddings of the entire graph $\mathbf{N_t}$ and the encoded instruction \textbf{W} are applied to a multi-layer cross-modal transformer as shown in Fig. \ref{fig:StratXplore}. The cross-attention and graph-aware self-attention (GASA) \cite{Chen2022DUET} in the transformer, model the language-viewpoint inter-modal relationships. GASA considers node embeddings and viewpoint adjacency to calculate global node-instruction embedding, $\mathbf{G_t}$. This is used for the exploitation action proposal.  

\subsubsection{Ego-centric Semantic Map Encoding}
\label{sec:ego_enc}
Local planning requires knowing which direction is to be pursued next. For this, we devise the local action space as follows. At first, the semantic features of the viewpoint are obtained using an object detector. Then, the ego-centric (polar) semantic-metric map is obtained by inverse-projecting the semantic features from the image space to the world space using the viewpoint's depth map. Technically, this is done by shooting rays from the camera centre to the semantic feature using the depth value of each pixel (lift) and projecting each feature onto the world ground plane (splat) following \cite{An2022BEVBert}. Each cell of the Birds-Eye-View (BEV) map represents a region of ground plane and contains average-pooled semantic features of that area. Finally, the map encoding $\mathbf{M_t}$ is obtained as the cell-wise sum of the polar feature map, navigability features (signifying occlusions/obstructions) and polar position embeddings of the grid. To capture instruction-viewpoint alignment useful for local decision making, we use a cross-modal (local) transformer with cross and self-attention and obtain the cell-instruction contextual representation, $\mathbf{C_t}$.

\subsection{Action Proposal}
\label{sec:act_prop}
For the exploitation action proposal, we follow previous methods \cite{Chen2022DUET,An2022BEVBert} for fusing the global and local contextual embeddings from the respective transformers (CMT) to obtain the global action proposal. At each step $t$, the action scores for each viewpoints are obtained as,  
\begin{align}
    \label{eq:trans}
    \mathbf{G_t} &= \mathrm{CMT}_{global}(\mathbf{W},\mathbf{N_t})\\
    s_g &= \mathrm{FFN}_g(\mathbf{G_t})\\
    \mathbf{C_t} &= \mathrm{CMT}_{local}(\mathbf{W},\mathbf{M_t})\\
    s_l &= \mathrm{FFN}(\mathbf{C_t})\\
    a_t &= arg \max~p([s_g;s_l] \mathbf{W})
\end{align}

where FFNs are feed forward neural networks that predict action scores for global and local contexts. The action proposal selects the node with the highest exploit action probability $a_t$ after fusing (represented by $[;]$) the global $s_g$ and local $s_l$ action features. 

This proposal is prone to navigational mistakes due to task complexity and agent needs to identify and recover from it. Because the network weights are not shared between the both CMTs, contextual representation generated by one does not affect the other. Hence, the object landmark detections are localised to the local map encoder and place landmarks are localised to global planner. This hinders effective error correction. To alleviate this, StratXplore considers both global and local landmark information for recovery.

\subsubsection{Detecting a navigation mistake}
The agent needs to have an implicit notion of navigational progress for successful navigation. For this, we propose two training schemes aimed at imparting deviation awareness and progress monitoring abilities to the planner. Firstly, we propose offline pre-training of the global cross-modal transformer (\S\ref{sec:pretraining}) to detect agent deviating from the optimal path (Fig. \ref{fig:StratXplore} (b.1)). Secondly, we use a predictor to estimate a \textit{confidence score} $S^{conf}$ during navigation. If the score for each of the candidate directions is less than a threshold $c_{thresh}$, the agent chooses to explore, otherwise it continues to exploit. 
Unlike existing self-monitoring agents which estimate the navigation progress by training an neural network to predict the distance to goal location, this module estimates the likelihood of recovering to the optimal path for a candidate viewpoint chosen by the agent. This is a fine-grained mistake estimation signal predicted from the cell-instruction embedding $\mathbf{C_t}$ as follows
 \begin{equation}
    \label{eq:conf}
     S^{conf} = sigmoid(\mathrm{FFN}_{c}(\mathbf{C_t}))
 \end{equation}
 
 The training process is explained in detail in \S\ref{sec:rcr-train}.

\subsubsection{Recovery}
We hypothesise that the local action scores, novelty and task-conformity (instruction-viewpoint correspondence) assigned to all frontiers are important for assessing \textit{relevant} frontiers for recovery. For instance, consider a frontier node, observed from various neighbouring viewpoints, obtains the relatively high action score from these observations. Logically, this direction (or frontier) is a good candidate as a \textit{relevant} frontier based on its cumulative action score assigned from independent observations. In addition, exploring novel frontiers can yield unique information about the environment. These aspects enhance the agent's ability to identify the optimal frontier. Four scores are used to rank the candidates, namely, action proposal score $S^{act}$, the novelty of the viewpoint $S^{novel}$, the alignment of the viewpoint to the instruction $S^{align}$, and the recency of the viewpoint $S^{recency}$. We explain them in detail in the following section. 


\section{Action-and-Knowledge based Frontier Selection}
\label{sec:akfs}
In this section, we explain the implementation of path correction and frontier selection. A frontier has two-levels of \textit{relevance} -  global and local. The local relevance is same as the action scores allocated by the Action Proposal module during the exploitation phase. This ensures that the scores relevant to any viewpoint at the time it is observed, are used for decision making. Global relevance is governed by temporal-recency to the current node as well as task-conformity. In effect, during frontier selection, local and global relevance are together considered for frontier ranking. To realise this, we use two memories namely: an Action memory (local relevance) and a Scene-Object memory dealing with the task-conformity and novelty (global relevance).

\subsection{Action Memory}
\label{sec:act_memory}
The action memory is a directed graph $G_{act} = <V_{t_{obs}} \epsilon_{score}>$ where $V_{t_{obs}}$ represents both visited viewpoints and frontiers and $\epsilon_{score}$ represents viewpoint adjacency and their action score proposed by the planner. Each $V_{t_{obs}}$ includes three attributes: a viewpoint identifier, the latest observation time step $t_{obs}$ and a visitation flag. The time step represents the order of visitation or observation i.e. both the visited viewpoint and its observed neighbours have the same time step $t=t_{obs}$. The $flag$ indicates if the node has been \verb|visited| or not (i.e. \verb|frontier|). The edges $\epsilon_{score}$ store the action score predicted by the action proposal module at each step. During exploitation, the edges that connect the visited nodes are set to 0, to prevent re-visitation. Note that action scores are normalised and can be considered as probabilities only in the neighbourhood but not in the overall graph context. Action scoring can be summarised as follows.

\label{sec:act_score}
 At time step $t$, the agent observes the environment and obtains a viewpoint, its neighbours and their connectivity. The current node obtains ($flag=$\verb|visited|, $t_{obs}=t$)  and the neighbours obtain ($flag=$\verb|frontier|, $t_{obs}=t$). The edges $\epsilon_{score}$ of the neighbours are updated based on the action scores predicted by the action proposal module.

\subsection{Scene-object Memory}
\label{sec:so_memory}

Our scene-object memory provides additional viewpoint information to the planner in order to compare frontiers and make global decisions. Knowledge of relevant objects in a viewpoint assists the agent in strictly complying with the instruction. It is also useful for selecting novel frontiers that are different from the visited locations to prevent repeated actions. We develop a scene-object memory for this purpose.

\begin{figure}
    \centering
    \begin{overpic}[width=0.8\columnwidth,percent]{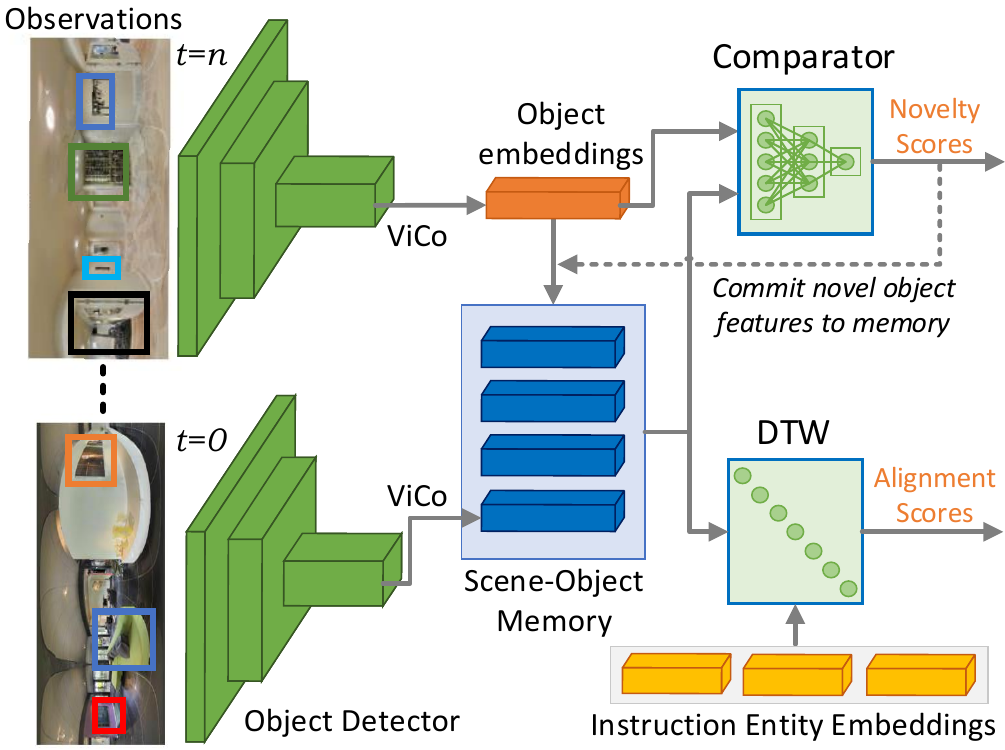}
    \put(45,18.5){\hyperref[sec:so_memory]{\phantom{\rule{42pt}{55pt}}}}
    \put(71,51){\hyperref[eq:novel]{\phantom{\rule{42pt}{40pt}}}}
    \put(71,15){\hyperref[eq:align]{\phantom{\rule{42pt}{40pt}}}}
    
    \end{overpic}
    \caption{\textbf{Novelty and Alignment Scoring.} At each time step, object features are added to memory if the objects are novel. During exploration both scores are used to rank frontiers.}
    \label{fig:so_memory}
    \vspace{-1.5em}
\end{figure}

The Scene-Object memory (Fig. \ref{fig:so_memory}) is a buffer accumulating object-related knowledge from viewpoints. This knowledge vector of a viewpoint is added to the memory only if it is novel (i.e. represents unique objects) with respect to other viewpoint knowledge vectors in the memory.
Note that here the measure of uniqueness is object-centric, and hence high similarity between knowledge vectors of two viewpoints in the memory means they have more or less the same type of objects. 

The knowledge vector is constructed for each viewpoint as follows. For each direction of the viewpoint, we select the top-K high confidence objects detected by an object detection model. We use the Faster R-CNN model \cite{Ren2017FasterRCNN} specifically trained on the Visual Genome dataset \cite{krishna2017visualgenome}. To represent the objects' \textit{knowledge} for a viewpoint, the ViCo \cite{Gupta2019ViCo} word embeddings of the object names are used. ViCo embeddings have visual and textual co-occurrence awareness i.e. objects that are seen together in the real world (chair and table) as well as in text (like in GloVe embeddings), are close in embedding space. In order to summarise the viewpoint objects, the sum of all object embeddings are used. As the frontiers are only partially observed from different angles via candidate directions of visited viewpoints, the frontier knowledge is the sum of object features of the respective directions. Contrastingly, the knowledge of the visited viewpoint $K_{v_i}$ is the sum of object features of non-candidate directions. Next, we explain our method for measuring novelty and entity alignment.
\subsection{Frontier Scoring}
\subsubsection{Novelty Scoring}
\label{sec:novel_score}

At each time step $t$, the navigation graph is updated with unique knowledge of the node at that step. For the current node, the novelty score is the inverse of the cosine similarity between the viewpoint knowledge and all the elements in the memory:

\begin{equation}
\label{eq:novel}
S^{novel}_{\mathbf{O}_t} = \frac{\left\Vert K_{\mathbf{O}_t} \right\Vert \left\Vert \sum K^{mem \setminus \mathbf{O}_t} \right\Vert}{(K_{\mathbf{O}_t} \cdot \sum K^{mem \setminus \mathbf{O}_t})}   
\end{equation}

The knowledge is added to memory if  $S^{novel}_{\mathbf{O}_t} > 2$, to signify largely different viewpoint objects. 

\subsubsection{Entity alignment scoring}
\label{sec:ea_score}

Unlike the language-vision correspondence measured by the cross-modal transformer, this alignment score considers landmark words or entities (rooms, objects etc.) extracted from the instruction and compares it against the elements in the memory (Fig. \ref{fig:priority}). To ensure the unexplored viewpoint monotonically aligns with the instruction, we measure the Dynamic Time Warping (DTW) \cite{Ilharco2019DTW} cost between the entity sequence from the instruction and the knowledge vectors corresponding to viewpoints of the test path. A test path is a sequence of viewpoints that leads to a frontier. First to compute DTW cost, ViCo embedded entities are extracted from the instruction $K^{entities} = ViCo($\verb|entities|$)$ and the knowledge vectors are extracted from the Scene-Object memory corresponding to test path. To reduce the computation cost in case of a large number of test paths, we filter frontiers with $S^{act} > 0.5$  and construct the test path as the shortest path $T_{\mathbf{O}_i}$ from the earliest visited viewpoints in the agent's trajectory, to every frontier viewpoint. The DTW score between  the $K^{entities}$ and the knowledge sequence of the test paths in $T_{\mathbf{O}_i}$ (eg. the path 1234H in the figure). We use a computationally less expensive but accurate implementation for small sequences, FastDTW \cite{Salvador2004FastDTWTA}, with the euclidean distance as the distance function. The DTW cost is normalised to obtain the alignment score:
\begin{equation}
\label{eq:align}
S^{align}_{\mathbf{O}_t} = exp\bigg(-\frac{DTW(K^{T}_{\mathbf{O}_i},K^{entities})}{|K^{T}_{\mathbf{O}_i}||K^{entities}|}\bigg)    
\end{equation}

\subsection{Frontier Selection}
\label{sec:frontierselection}
\begin{figure}
    \centering
    \includegraphics[width=\columnwidth]{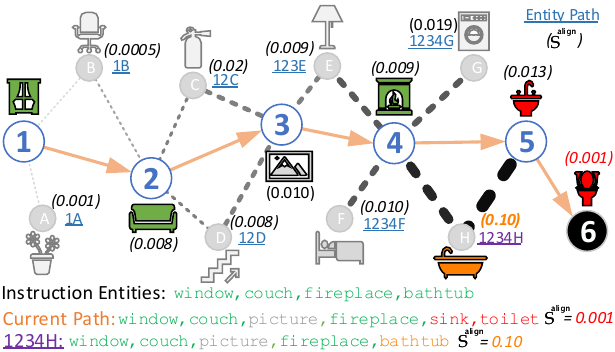}
    \caption{\textbf{Example for Temporal prioritisation and Entity Alignment Scoring.} Recent memories (thicker connections) are prioritised over past actions to encourage exploration. $S^{align}$ (values in parenthesis) is the highest for the path to the optimal frontier (here \textcolor{violet}{\underline{1234H}}).}
    \label{fig:priority}
    \vspace{-1.5em}
\end{figure}
In the exploration mode, the memory is queried to find the optimal frontier. For this, the frontier selector queries the memory with all the un-normalised action scores. We rank the nodes as follows:
\begin{enumerate}
    \item Select frontier nodes from action memory i.e. $flag=$ \verb|frontier| and $t_{obs} < t$. 
    \item Score accumulation: Action scores assigned to incident edges of each frontier node $\epsilon_{score}$ are summed to obtain $S^{act}$ of that frontier. This ensures frontiers that are relevant locally, is also relevant globally.
    \item Temporal relevance calculation: We obtain the recency score $S^{recency} =  \exp(\gamma(t_i-t))$ where the scores of nodes from recent history remain the same while scores from early steps are diminished based on decay factor $\gamma$ (Fig. \ref{fig:priority}).
    \item Obtain the novelty score $S^{novel}$ for each unexplored location with respect to the current viewpoint.
    \item Final scores for frontiers are obtained as $ S = S^{act} * S^{recency}(S^{novel} + S^{align})$. Scores are normalised.
\end{enumerate}

The node with the maximum score after normalisation is the optimal frontier. The agent executes the shortest path through the navigation graph to reach the node.

\section{Training}
\subsection{Pre-training}

\label{sec:pretraining}
Previous studies have applied different pre-training tasks to improve task generalisation in VLN. Pre-training provides a holistic understanding of the task to the cross-modal planner and is a good starting point for downstream navigation models \cite{An2022BEVBert,Hao2020PREVALENT}. Accordingly, we begin by pre-training our model with behaviour cloning based on offline expert demonstrations, alongside various vision-and-language related tasks. These tasks include masked language modelling (MLM), masked region classification (MRC), single-step action prediction (SAP), and object grounding (OG).

In order to train cross-modal transformers to identify deviation from the ground truth path, we introduce a deviation prediction (DP) task that predicts if parts of the agent trajectory sequence have deviated from the instruction path. The global node-language embedding $\mathbf{G_t}$ is fed to a 2 layer FFN and trained together with the aforementioned auxiliary tasks. This auxiliary training requires synthetic path demonstrations, which are derived from the R2R dataset. For this, we introduce carefully controlled perturbations to the ground truth paths. The perturbed path is comprised of parts of the ground truth path and a detour path commencing from a random viewpoint. The detour may end at: a location previously traversed in the ground truth path, one of the frontiers, or a random location in the vicinity (but not neighbours) of the ground truth path. The detoured segment is reconnected back with the remainder of the ground truth path via the shortest path.

The viewpoints of the resulting path of length $N$ are labelled as on track (0), deviated (1), or recovering (2) i.e. ${lab_i}_1^N \in \{0, 1, 2\}$. The classifier is optimised using cross entropy loss.
\subsection{Navigation Training}
We train the agent using imitation learning (IL). In IL, a teacher model suggests the next action based on the ground truth path and the action prediction is optimised using cross entropy loss. 

\subsubsection{Recovery Confidence Prediction}
\label{sec:rcr-train}
The recovery confidence prediction model is trained online using teacher-forcing. The nearest ground truth viewpoint for each candidate direction of the current viewpoint is calculated during navigation training.  The confidence score, predicted by the model  $p^{rec}_{c,t}$ for each candidate $c$, signifies the likelihood of recovery. The training target $y^{rec}_{c,t}$ is the normalised distance $d_{c}$ from each candidate viewpoint $c$ to the nearest location in the ground truth path i.e. $d_c = min(d_{c,GT}) / d_{max}$. The target will be 1 if the agent is already on the ground truth path and $(1 - d_c)$ as it deviates from it. The objective is to minimise the mean squared error (MSE) between the target and predicted confidence scores, 
\begin{equation}
    \mathcal{L}_{rcr} = \sum_{t=1}^T (y^{rec}_{c,t} - p^{rec}_{c,t})^2
\end{equation}

\subsubsection{Action prediction}
The action selection objective is optimised with a cross-entropy loss. The overall loss is the weighed sum of action prediction and recovery confidence, 

\begin{equation}
    \mathcal{L}_{loss} = - \lambda \sum_{t=1}^T (y^{nv}_{c,t} \log(p_{c,t}) -  (1 - \lambda) \mathcal{L}_{rcr}
\end{equation}

where $p_{c,t}$ is the action probability of thecandidate direction $c$ in viewpoint at step $t$, $y_{c,t}^{nv} \in \{0, 1\}$ indicates ground-truth action, $\lambda = 0.4$ is the weight balancing the two losses.

\section{Experiments}
We evaluate StarXplore using two VLN datasets for testing its room-finding capabilities, namely Room-to-Room (R2R) \cite{Anderson2018R2R}, Room-for-Room (R4R) and \cite{Jain2020R4R}. R2R has short turn-by-turn instructions, and R4R has coarser instructions and longer trajectories extended from R2R.
\subsection{Implementation Details}
We adopt BEVBert \cite{An2022BEVBert}, a topology-learning transformer navigator that uses Birds-Eye-View map for local action prediction, as our baseline. We extend this model with our deviation pretraining, recovery confidence prediction and frontier selection, to impart error recovery capability. The auxiliary tasks (\S\ref{sec:pretraining}) used to pre-train the CMTs are mixed using the ratio MLM:SAP:MRC:OG:DP = 5:5:1:1:1:1. The hyperparameters for the baseline are set according to the model published in \cite{An2022BEVBert}. 

The action score memory has the temporal priority factor $\gamma$ set to 0.1. The local BEV map size is set to 21 grids with 0.5m resolution based on the recommendation from the original work.

\subsection{Evaluation Metrics}
For evaluating the model's performance on R2R \cite{Anderson2018R2R}, four widely recognised metrics are employed: Trajectory Length (TL), Navigation Error (NE), Success Rate (SR) and Success Rate weighted by Path Length (SPL). NE is the distance from the goal to the agent's stopping position and SR assesses the frequency of successfully reaching the goal within 3m. 

Additionally, in R4R \cite{Jain2020R4R}, three other metrics are adopted as per existing studies: Coverage weighted by Length Score (CLS), Normalised Dynamic Time Warping (nDTW) \cite{Ilharco2019DTW}, and Success rate weighted by normalized Dynamic Time Warping (SDTW), further expanding the evaluation framework. 

\section{Results}
\label{sec:results}

\begin{table*}[ht]
\centering
\caption{\small{Qualitative comparison (\S\ref{sec:r2rresult}) with state-of-the-art methods on R2R dataset.}\vspace{-.3cm}}
\begin{tabular}{l|cccc|cccc|cccc}
\toprule
\textbf{Methods} & \multicolumn{4}{c|}{\textbf{Val Seen}} & \multicolumn{4}{c|}{\textbf{Val Unseen}} & \multicolumn{4}{c}{\textbf{Test Unseen}} \\
 &  SR$\uparrow$ & SPL$\uparrow$ & TL & NE$\downarrow$ & SR$\uparrow$ & SPL$\uparrow$ & TL & NE$\downarrow$ & SR$\uparrow$ & SPL$\uparrow$ & TL & NE$\downarrow$ \\ \hline\hline
Random & 16 & - & 9.58 & 9.45 & 16 & - & - & 9.23 & 13 & 12 & 9.89 & 9.79 \\
Human \cite{Fried2018SF}  & - & - & - & - & - & - & - & - & 86 & 76 & 11.90 & 1.16 \\ \midrule
Seq2Seq~\cite{Anderson2018R2R} & 6.0 & 39 & 11.33 & - & 22 & - & \textbf{8.39} & 7.84 & 20 & 18 & \textbf{8.13} & 7.85 \\

VLN$\circlearrowright$BERT~\cite{Hong2021RecVLNBERT} & 72 & 68 & \textbf{11.13} & 2.90 & 63 & 57 & 12.01 & 3.93 & 63 & 57 & 12.35 & 4.09 \\
Self-monitoring$^\dagger$~\cite{Ma2019SelfMonitoring} & 69 & 63 & 11.69 & 3.31 & 47 & 41 & 12.61 & 5.48 & 61 & 56 & - & 4.48 \\
Regretful-Agent~\cite{Ma2019Regretful} & 69 & 63 & - & 3.23 & 50 & 41 & - & 5.32 & 48 & 40 & - & 5.69 \\
AuxRN \cite{Zhu2020AuxRN} & 70 & 67 & - & 3.33  & 55 & 50 & - & 5.28 & 55 & 51 & - & 5.15 \\
HAMT~\cite{Chen2021HAMT} &  76 & 72 & 11.15 & 2.51 &  66 & 61 & 11.46 & \textbf{2.29} & 65 & 60 & 12.27 & 3.93 \\
SSM~\cite{Wang2021SSM} & 71 & 62 & 14.7 & 3.10 & 62 & 45 & 20.7 & 4.32 & 61 & 46 & 20.4 & 4.57 \\ 
DUET~\cite{Chen2022DUET} &  79 & 73 & 12.32 & 2.28 & 72 & 60 & 13.94 & 3.31 & 69 & 59 & 14.73 & 3.65 \\
BEVBert~\cite{An2022BEVBert} & - & - & - & - & 75 & 64 & - & 2.81 & 73 & 62 & - & 3.13 \\ 
\midrule
\textbf{StratXplore} (Ours) &\textbf{80.2}&\textbf{75.4}&12.16&\textbf{2.47}&\textbf{77.61}&\textbf{66.92}&12.94&\textbf{2.93}&\textbf{75.86}&\textbf{64.79}&14.36&\textbf{3.04} \\
\bottomrule

\end{tabular}
\label{tab:r2rresult}

\end{table*}
\begin{table*}[ht]
    \centering

    \caption{\small{Qualitative comparison (\S\ref{sec:r4rresult}) with the state-of-the-art methods on  R4R  dataset.}\vspace{-.3cm}} 
    \begin{tabular}{l|cccccc|cccccc}
    \toprule 
    ~ & \multicolumn{6}{c|}{\textbf{Val Seen}}& \multicolumn{6}{c}{\textbf{Val Unseen}} \\
    \multirow{-3}{*}{\textbf{Methods}} &NE$\downarrow$ &TL$\downarrow$ &SR$\uparrow$ & CLS$\uparrow$ & nDTW$\uparrow$ & SDTW$\uparrow$ &NE$\downarrow$ &TL$\downarrow$ &SR$\uparrow$ & CLS$\uparrow$ & nDTW$\uparrow$ & SDTW$\uparrow$\\
    \hline
    \midrule
    Speaker-Follower~\cite{Fried2018SF} & 5.35 & 15.4  & 52 & 0.46 & - & - & 8.47 & 19.9  & 24 & 0.30 & - & - \\
    RCM~\cite{Wang2019RCM} & 5.37 & 18.8  & 53 & 0.55 & - & - & 8.08 & 28.5  & 26 & 0.35 & 0.30 & 0.13\\
    PTA (high-level)~\cite{landi2021PTA} & \textbf{4.54} & 16.5  & 58 & 0.60 & 0.58 & 0.41 & 8.25 & \textbf{17.7}  & 24 & 0.37 & 0.32 & 0.10\\
    EGP~\cite{Deng2020EGP} & - & -  & - & - & - & - & \textbf{8.00} & 18.3  & 30 & 0.44 & 0.37 & 0.18\\
    E-Drop~\cite{Tan2019EnvDrop} & - & 19.9  & 52 & 0.53 & - & 0.27 & - & 27.0  & 29 & 0.34 & - & 0.09 \\
    OAAM~\cite{Qi2020OAAM} & - & \textbf{11.8}  & 56 & 0.54 & - & 0.32 & - & 13.8  & 31 & 0.40 & -  & 0.11\\
    BabyWalk~\cite{Zhu2020BabyWalk} & - & - & - & - & - & - &  8.20 & 19.0 & 27 & 0.49 & 0.39 & 0.18 \\
    EntityGraph~\cite{Hong2020RelGraph} & 5.31 & - & 52 & 0.55 & 0.62 & 0.50 & 7.43 & - & 36 & 0.41 & \textbf{0.47} & \textbf{0.34}\\
    SSM \cite{Wang2021SSM}&4.60 &19.4 & 63 & 0.65 & 0.56 &0.44&8.27 &22.1 &32 &0.53 &0.39 &0.19\\
    \midrule
    \textbf{StratXplore} (Ours) & \textbf{5.26}&20.35&\textbf{66} &\textbf{0.67}&0.58&\textbf{0.46}&8.10&21.64&\textbf{38}&\textbf{0.55}&0.45&0.17\\
    \bottomrule
    \end{tabular}
        \vspace*{-4pt}
   \label{table:r4rresult}
    \vspace*{-7pt}
    \end{table*}
\subsection{R2R dataset}
\label{sec:r2rresult}

Table \ref{tab:r2rresult} compares the performance of StratXplore with current methods on the R2R task. The results displayed in the last row demonstrate that our model outperforms existing methods on SR and SPL across dataset split. Compared to our baseline, BEVBert \cite{An2022BEVBert}, we can observe that the proposed model obtains relative improvement in SR (3.91\%) and SPL (2.79\%) in Test Unseen split. In particular, our method outperforms other explicit memory models such as SSM \cite{Wang2021SSM} (SR: +14.86, SPL: +18.79) and DUET \cite{Chen2022DUET} (SR: +6.89, SPL: +5.79) by an absolute margin. Similar improvement is seen when compared to progress monitoring methods such as \cite{Ma2019SelfMonitoring,Zhu2020AuxRN,Ma2019Regretful}, underlining the effectiveness of our proposed method.

\subsection{R4R dataset}
\label{sec:r4rresult}

Table \ref{table:r4rresult} compares our model to other VLN methods on the R4R task. Our model shows the best success rate in both ValSeen (66\%) and ValUnseen (35\%) splits. StratXplore shows a relative improvement of 3. 17\% and 3. 07\% on SR and CLS in Val Seen split compared to SSM, respectively. Similarly, we see a relative improvement of 5. 55\% (SR) and 3. 77\% (CLS) on the Val Unseen split. This clearly demonstrates the impact of our method on long-horizon path planning tasks. However, the path correction has adversely impacted the TL and trajectory-shape-dependent scores such as nDTW and sDTW, nonetheless, the results are still comparable to those of the prior agents. 

\subsection{Qualitative comparison}
\begin{figure}
    \centering
    \includegraphics[width=\columnwidth]{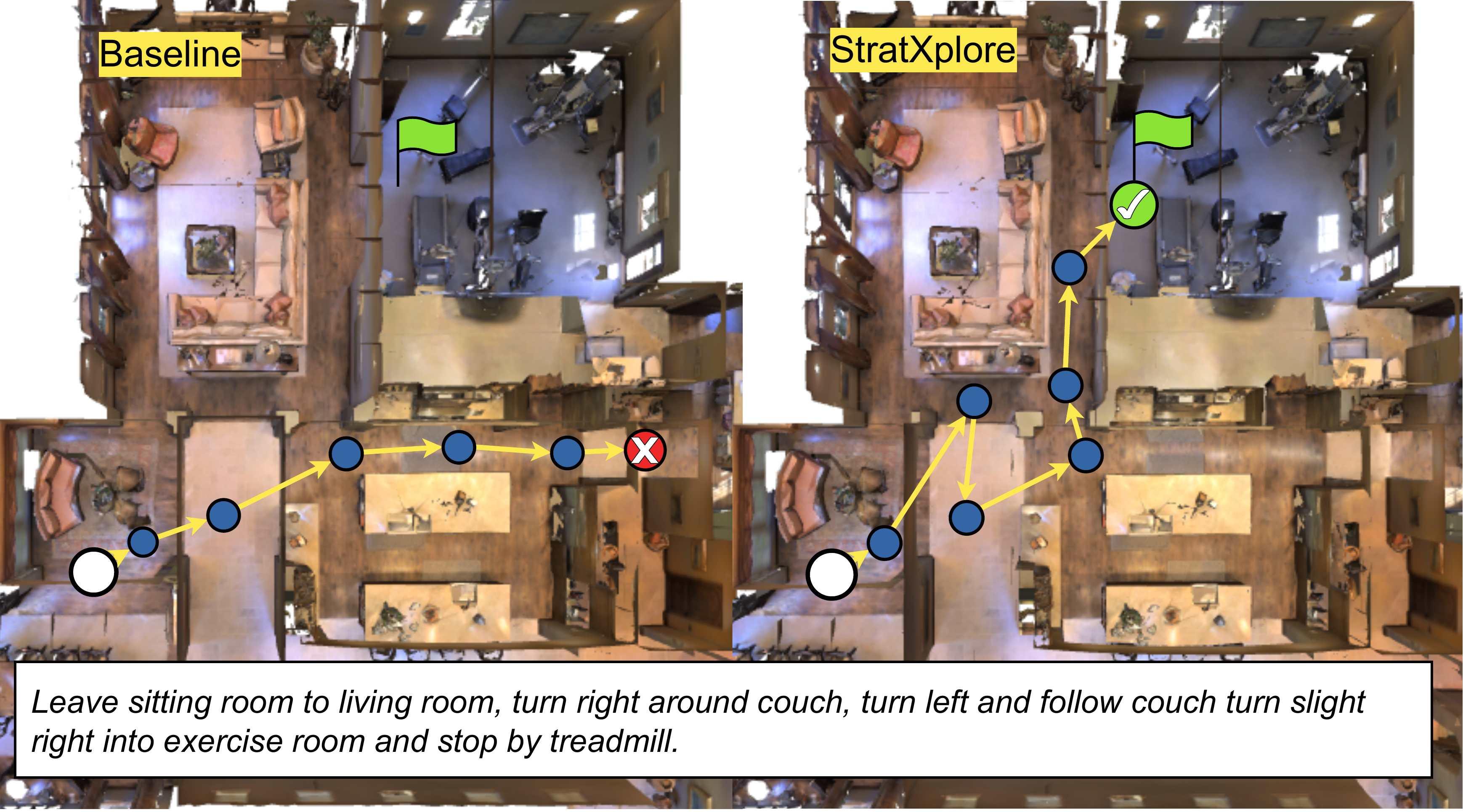}
    \caption{\textbf{Qualitative comparison between trajectories of the baseline and StratXplore agents.} The baseline agent does not recover from the navigational mistake and continues to exploit the same direction and eventually fails. Although StratXplore makes a mistake by entering the second living room, it quickly corrects itself by moving to the best observed frontier (hallway).}
    \label{fig:qual-comp}
        \vspace{-1.5em}
\end{figure}
We sample an interesting scenario from R2R ValUnseen split (Fig. \ref{fig:qual-comp}). While both the baseline and our StratXplore agent make mistakes, StratXplore identifies the navigational mistake and recovers to the optimal frontier while the baseline agent fails to do so. Interestingly, deviation prediction training helps the agent identify the correct left turn.
\subsection{Perturbation Study}
\begin{table}[ht]
\centering
\caption{\small{Change in  success rates after Kidnapping agents with different memory types}\vspace{-.3cm}} 
\resizebox{\columnwidth}{!}{\fontsize{12}{13}\selectfont
\begin{tabular}{l|c|cccc}
\toprule

\multirow{2}{*}{\textbf{Memory}} & \multirow{2}{*}{\begin{tabular}[c]{@{}c@{}}\textbf{Baseline}\\ (SR,SPL)\end{tabular}} & \multicolumn{4}{c}{\textbf{Change in Success Rate ($\Delta$SR,$\Delta$SPL)}} \\ 
                                &                              & Visited & Guiding & 3-Neighbourhood & Close   \\ \hline
    \midrule
\#1 Recurrent & (63,57) & -9,-17   & -6,-11   & -12,-17       & -30,-36  \\
\#2 Hierarchical & (66,61) & -9,-14   & -7,-11   & -8,-10        & -24,-30  \\
\#3 Topological & (72,60) & -7,-10   & +2,+1   & -3,-7         & -22,-25 \\
\#4 Topo-metric  & (75,64) & -5,-7   & +1,+1   & -4,-4         & -18,-21 \\
\#5 Ours & (76,65) & -4,-6   & +3,+2     & -2,-3         & -13,-17 \\ 
\bottomrule
\end{tabular}
}
\label{tab:kidnapping}
\end{table}
We study the agent's ability to recover by \textit{kidnapping} it to various viewpoints in the environment and measuring the change in navigational success (Table \ref{tab:kidnapping}). For this we devise 4 kidnapping scenarios with an increasing order of difficulty, i.e. kidnapping: 1) to a previously visited location (Visited), 2) to a location from the ground truth trajectory (Guiding), 3) toa 3-hop neighbourhood of the current path (3-Neighbourhood), and 4) to a random location close to the trajectory (Close). The first two scenarios measure the instruction-trajectory co-grounding ability while the latter two measure recover-ability of the planner during a critical failure. We test the performance of different memory types of the respective representative models: Recurrent (VLN$\circlearrowright$BERT \cite{Hong2021RVLNBERT}), Hierarchical (VLN-HAMT \cite{Chen2021HAMT}), Topological (DUET \cite{Chen2022DUET}), Topo-metric (BEVBert \cite{An2022BEVBert}) and Dual Action-Novelty (Ours). The recurrent memory has the largest drop in SR (-9,-6,-12,-30)  and SPL  (-17,-11,-17,-36) in all kidnapping scenarios, revealing the inefficacy of short term memories in path recovery. Topological and Topo-metric memories demonstrate better recovery compared to methods \#1 and \#2.  It is interesting to see that only the topological memories and our methods (Ours) leveraged guidance towards the goal. Also, in methods \#1-\#4 the success rates drop considerably even for visited viewpoints. One explanation is that the agent continues to repeat previous actions and fails to recover toward the goal. In contrast, our method identifies novel environments to navigate and recovers to novel frontiers in recent history. This allows the agent to identify the location it is kidnapped to and continue the navigation from one of the candidate frontiers.  Hence, StratXplore demonstrates the lowest change in success rates ((4,-6),(+3,+2),(-2,-3),(-13,-17)) among all existing memory types used in VLN and can also leverage the movement towards the goal.

\section{Conclusion}
In this paper, we introduce a strategic exploration model, called StratXplore, designed to tackle the challenges encountered by Vision-Language-Navigation agents. We recognise error recovery as an essential capability of an embodied robot navigating in unseen or novel environments. Our method imparts four important aspects to VLN agents for error recovery: progress monitoring, ensuring task-conformity, seeking novel viewpoints and identifying a viewpoint's temporal-importance. Experimental results on R2R and R4R tasks demonstrate that our method is effective in improving navigational success in unseen environments. 

\noindent\textbf{Limitations and future work} StratXplore agent rely on post-facto comparison of frontiers after a navigation mistake. To improve implicit awareness of a mistake and reduce added ranking cost, this method could be integrated with the planner rather than a separate error recovery module. This will be addressed in future work. 

\bibliographystyle{IEEEtran}
\bibliography{IEEEabrv,bibliography}

\end{document}